\documentclass[11pt,a4paper]{article}
\usepackage[hyperref]{emnlp2020}
\usepackage{times}
\usepackage{latexsym}
\usepackage{amsmath}
\usepackage{amssymb}
\usepackage[T1]{fontenc}
\usepackage{todonotes}
\usepackage[normalem]{ulem}
\usepackage{booktabs}
\usepackage{arydshln}
\usepackage{pifont}
\usepackage{multirow}
\usepackage{xspace}

\usepackage{microtype}

\aclfinalcopy 
\def\aclpaperid{2470} 

\newcommand{\cls}{\texttt{[CLS]}\xspace}
\newcommand{\sep}{\texttt{[SEP]}\xspace}

\newcommand{\updoc}{\mathrm{doc}}
\newcommand{\upsent}{\mathrm{sent}}
\newcommand{\upspan}{\mathrm{span}}

\newcommand{\ext}{CUPS$_\textrm{EXT}$\xspace}
\newcommand{\cmp}{CUPS$_\textrm{CMP}$\xspace}

\newcommand{\modelext}{\texttt{ext}\xspace}
\newcommand{\modelabs}{\texttt{abs}\xspace}
\newcommand{\modelcmp}{\texttt{cmp}\xspace}

\definecolor{delete}{RGB}{165,0,3}
\newcommand{\compress}[1]{\textcolor{delete}{\sout{#1}}}

\title{Compressive Summarization with Plausibility and Salience Modeling}

\author{Shrey Desai \quad\quad Jiacheng Xu \quad\quad Greg Durrett \\
  Department of Computer Science \\
  The University of Texas at Austin \\
  \texttt{shreydesai@utexas.edu\quad\{jcxu, gdurrett\}@cs.utexas.edu}}

\date{}

\begin{document}
\maketitle

\begin{abstract}
Compressive summarization systems typically rely on a crafted set of syntactic rules to determine what spans of possible summary sentences can be deleted, then learn a model of what to actually delete by optimizing for content selection (ROUGE). In this work, we propose to relax the rigid syntactic constraints on candidate spans and instead leave compression decisions to two data-driven criteria: plausibility and salience. Deleting a span is \textit{plausible} if removing it maintains the grammaticality and factuality of a sentence, and spans are \textit{salient} if they contain important information from the summary. Each of these is judged by a pre-trained Transformer model, and only deletions that are both plausible and not salient can be applied. When integrated into a simple extraction-compression pipeline, our method achieves strong in-domain results on benchmark summarization datasets, and human evaluation shows that the plausibility model generally selects for grammatical and factual deletions. Furthermore, the flexibility of our approach allows it to generalize cross-domain: our system fine-tuned on only 500 samples from a new domain can match or exceed an in-domain extractive model trained on much more data.\footnote{Code and datasets available at \url{https://github.com/shreydesai/cups}}
\end{abstract}

\section{Introduction}

Compressive summarization systems offer an appealing tradeoff between the robustness of extractive models and the flexibility of abstractive models. Compression has historically been useful in heuristic-driven systems \cite{knight-marcu-2000-sentence-compression,knight-marcu-2002-sentence-compression,wang-etal-2013-a} or in systems with only certain components being learned \cite{martins-smith-2009-summarization,woodsend-lapata-2012-multi-aspect,qian-liu-2013-fast}. End-to-end learning-based compressive methods are not straightforward to train: exact derivations of which compressions should be applied are not available, and deriving oracles based on ROUGE \cite{berg-kirkpatrick-2011-jointly,durrett-etal-2016-learning,xu-durrett-2019-neural,mendes-etal-2019-jointly} optimizes only for content selection, not grammaticality or factuality of the summary. As a result, past approaches require significant engineering, such as creating a highly specific list of syntactic compression rules to identify permissible deletions \cite{berg-kirkpatrick-2011-jointly,li-etal-2014-improving-multi,wang-etal-2013-a,xu-durrett-2019-neural}. Such manually specified, hand-curated rules are fundamentally inflexible and hard to generalize to new domains.

In this work, we build a summarization system that compresses text in a more data-driven way. First, we create a small set of high-recall constituency-based compression rules that cover the space of legal deletions. Critically, these rules are merely used to propose candidate spans, and the ultimate deletion decisions are controlled by two data-driven models capturing different facets of the compression process. Specifically, we model \textit{plausibility} and \textit{salience} of span deletions. Plausibility is a domain-independent requirement that deletions maintain grammaticality and factuality, and salience is a domain-dependent notion that deletions should maximize content selection (from the standpoint of ROUGE). In order to learn plausibility, we leverage a pre-existing sentence compression dataset \cite{filippova-2013-overcoming}; our model learned from this data transfers well to the summarization settings we consider. Using these two models, we build a pipelined compressive system as follows: (1) an off-the-shelf extractive model highlights important sentences; (2) for each sentence, high-recall compression rules yield span candidates; (3) two pre-trained Transformer models \cite{clark-2020-electra} judge the plausibility and salience of spans, respectively, and only spans which are both plausible and not salient are deleted.

We evaluate our approach on several summarization benchmarks. On CNN \cite{hermann-2015-teaching}, WikiHow \cite{koupaee-wang-2018-wikihow}, XSum \cite{narayan-2018-xsum}, and Reddit \cite{kim-2019-reddit}, our compressive system consistently outperforms strong extractive methods by roughly 2 ROUGE-1, and on CNN/Daily Mail \cite{hermann-2015-teaching}, we achieve state-of-the-art ROUGE-1 by using our compression on top of MatchSum \cite{zhong-etal-2020-matchsum} extraction. We also perform additional analysis of each compression component: human evaluation shows plausibility generally yields grammatical and factual deletions, while salience is required to weigh the content relevance of plausible spans according to patterns learned during training.

Furthermore, we conduct out-of-domain experiments to examine the cross-domain generalizability of our approach. Because plausibility is a more domain-independent notion, we can hold our plausibility model constant and adapt the extraction and salience models to a new setting with a small number of examples. Our experiments consist of three transfer tasks, which mimic real-world domain shifts (e.g., newswire $\rightarrow$ social media). By fine-tuning salience with only 500 in-domain samples, we demonstrate our compressive system can match or exceed the ROUGE of an in-domain extractive model trained on tens of thousands of document-summary pairs.

\section{Plausible and Salient Compression}
\label{sec:compression-overview}

Our principal goal is to create a compressive summarization system that makes linguistically informed deletions in a way that generalizes cross-domain, without relying on heavily-engineered rules. In this section, we discuss our framework in detail and elaborate on the notions of plausibility and salience, two \textit{learnable} objectives that underlie our span-based compression.

\begin{figure}[t]
    \centering
    \includegraphics[width=6.5cm]{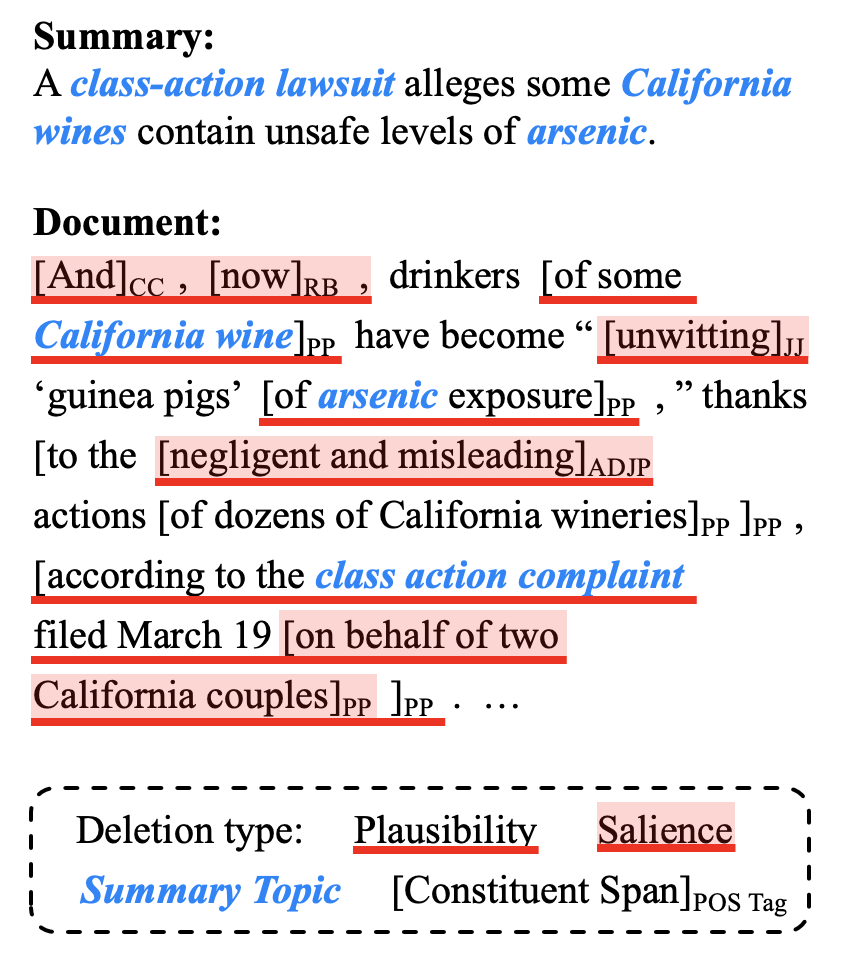}
    \caption{Decomposing span-based compression into plausibility and salience (\S\ref{sec:compression-overview}). Plausible compressions (underlined) must maintain grammaticality, thus [to the ...  wineries]$_\textrm{PP}$ is not a candidate. Salience identifies low-priority content from the perspective of this dataset (highlighted). Constituents both underlined \textit{and} highlighted are deleted.}
    \label{fig:compression-overview}
\end{figure}

\subsection{Plausibility}
\label{sec:plausibility}

\textbf{Plausible} compressions are those that, when applied, result in grammatical and factual sentences; that is, sentences that are syntactically permissible, linguistically acceptable to native speakers \cite{chomsky-1956-syntactic,schutze-1996-the}, and factually correct from the perspective of the original sentence. Satisfying these three criteria is challenging: acceptability is inherently subjective and measuring factuality in text generation is a major open problem \cite{kryscinski-etal-2020-factuality,wang-etal-2020-qags,durmus-etal-2020-feqa,goyal-durrett-2020-factuality}. Figure~\ref{fig:compression-overview} gives examples of plausible deletions: note that \textit{of dozens of California wineries} would be grammatical to delete but significantly impacts factuality.

We can learn this notion of plausibility in a data-driven way with appropriately labeled corpora. In particular, \newcite{filippova-2013-overcoming} construct a corpus from news headlines which can suit our purposes: these headlines preserve the important facts of the corresponding article sentence while omitting minor details, and they are written in an acceptable way. We can therefore leverage this type of supervision to learn a model that specifically identifies plausible deletions.

\subsection{Salience}
\label{sec:salience}

As we have described it, plausibility is a domain-\textit{independent} notion that asks if a compression maintains grammaticality and factuality. However, depending on the summarization task, a compressive system may not want to apply all plausible compressions. In Figure~\ref{fig:compression-overview}, for instance, deleting all plausible spans results in a loss of key information. In addition to plausibility, we use a domain-\textit{dependent} notion of \textbf{salience}, or whether a span should be included in summaries of the form we want to produce.

Labeled oracles for this notion of content relevance \cite[\textit{inter alia}]{gillick-favre-2009-a,berg-kirkpatrick-2011-jointly} can be derived from gold-standard summaries using ROUGE \cite{lin-2004-rouge}. We compare the ROUGE score of an extract with and without a particular span as a proxy for its importance, then learn a model to classify which spans improve ROUGE if deleted. By deleting spans which are both plausible and salient in Figure~\ref{fig:compression-overview}, we obtain a compressed sentence that captures core summary content with 28\% fewer tokens, while still being fully grammatical and factual.

\subsection{Syntactic Compression Rules}
\label{sec:syntactic-compression-rules}

The base set of spans which we judge for plausibility and salience comes from a recall-oriented set of compression rules over a constituency grammar; that is, they largely cover the space of valid deletions, but include invalid ones as well.

Our rules allow for deletion of the following: (1) parentheticals (PRN) and fragments (FRAG); (2) adjectives (JJ) and adjectival phrases (ADJP); (3) adverbs (RB) and adverbial phrases (ADVP); (4) prepositional phrases (PP); (5) appositive noun phrases (NP$_1$--[,--NP$_2$--,]); (6) relative clauses (SBAR); and (7) conjoined noun phrases (e.g., NP$_1$--[CC--NP$_2$]), verb phrases (e.g., VP$_1$--[CC--VP$_2$]), and sentences (e.g., S$_1$--[CC--S$_2$]). Brackets specify the constituent span(s) to be deleted, e.g., CC--NP$_2$ in NP$_1$--[CC--NP$_2$].

Much more refined rules would be needed to ensure grammaticality: for example, in \textit{She was [at the tennis courts]$_\textrm{PP}$}, deletion of the PP leads to an unacceptable sentence. However, this base set of spans is nevertheless a good set of building blocks, and reliance on syntax gives a useful inductive bias for generalization to other domains \cite{swayamdipta-etal-2018-syntactic}.

\section{Summarization System}
\label{sec:summarization-system}

We now describe our compressive summarization system that leverages our notions of plausibility and salience. For an input document, an off-the-shelf extractive model first chooses relevant sentences, then for each extracted sentence, our two compression models decide which sub-sentential spans to delete. Although the plausibility and salience models have different objectives, they both output a posterior over constituent spans, and thus use the same base model architecture.

We structure our model's decisions in terms of separate sentence extraction and compression decisions. Let $S_1,\ldots,S_n$ denote random variables for sentence extraction where $S_i = 1$ indicates that the $i$th sentence is selected to appear in the summary. Let $C_{11}^{\text{PL}},\ldots,C_{nm}^{\text{PL}}$, denote random variables for the plausibility model, where $C_{ij}^{\text{PL}} =1$ indicates that the $j$th span of the $i$th sentence is plausible. An analogous set of $C_{ij}^{\text{SAL}}$ is included for the salience model. These variables are modeled independently and fully specify a compressive summary; we describe this process more explicitly in Section~\ref{sec:inference}.

\subsection{Preprocessing}

Our system takes as input a document $D$ with sentences $s_1, \cdots, s_n$, where each sentence $s_i$ has words $w_{i1}, \cdots, w_{im}$. We constrain $n$ to be the maximum number of sentences that collectively have less than 512 wordpieces when tokenized. Each sentence has an associated constituency parse $T_i$ \cite{kitaev-klein-2018-constituency} comprised of constituents $c = (t, i', j')$ where $t$ is the constituent's part-of-speech tag and $(i', j')$ are the indices of the text span. Let $R(T_i)$ denote the set of spans proposed for deletion by our compression rules (see Section~\ref{sec:syntactic-compression-rules}).

\subsection{Extraction}

Our extraction model is a re-implementation of the BERTSum model \cite{liu-lapata-2019-text}, which predicts a set of sentences to select as an extractive summary. The model encodes the document sentences $s_1, \cdots, s_n$ using BERT \cite{devlin-etal-2019-bert}, also preprending \cls and adding \sep as a delimiter between sentences.\footnote{BERT can be replaced with other pre-trained encoders, such as ELECTRA \cite{clark-2020-electra}, which we use for most experiments.} We denote the token-level representations thus obtained as: $[\mathbf{h}^{\updoc}_{11}, \cdots, \mathbf{h}^{\updoc}_{nm}] = \mathrm{Encoder}([s_1, \cdots, s_n])$

During fine-tuning, the \cls tokens are treated as sentence-level representations. We collect the \cls vectors over all sentences $\mathbf{h}^{\updoc}_{i1}$, dot each with a weight vector $\mathbf{w} \in \mathbb{R}^{d}$, and use a sigmoid to obtain selection probabilities: $P(S_i=1|D) = \sigma(\mathbf{h}^{\updoc\top}_{i1} \mathbf{w})$

\begin{figure}[t]
    \centering
    \includegraphics[width=7.5cm]{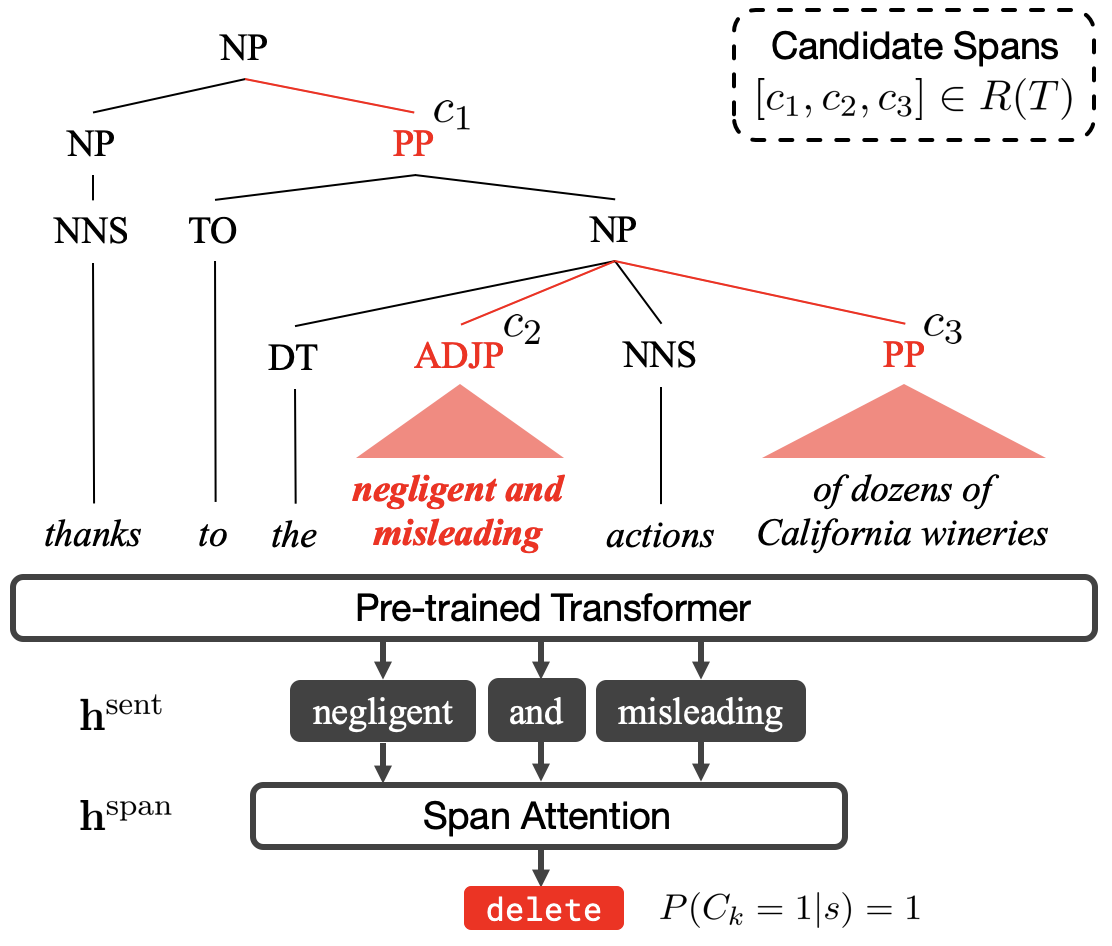}
    \caption{Compression model used for plausibility and salience modeling (\S\ref{sec:compression-model}). We extract candidate spans $c_i \in C(T)$ to delete, then compute span embeddings with pre-trained encoders (only one span embedding shown here). This embedding is then used to predict whether the span should be kept or deleted.}
    \label{fig:compression-example}
\end{figure}

\subsection{Compression}
\label{sec:compression-model}

Depicted in Figure~\ref{fig:compression-example}, the compression model (instantiated twice; once for plausibility and once for salience) is a sentence-level model that judges which constituent spans should be deleted. We encode a single sentence $s_i$ at a time, adding \cls and \sep as in the extraction model. We obtain token-level representations using a pre-trained Transformer encoder:\footnote{The encoders between the extraction and compression modules are fine-tuned separately; in other words, our modules do not share any parameters.} $[\mathbf{h}^{\upsent}_{i1}, \cdots, \mathbf{h}^{\upsent}_{im}] = \mathrm{Encoder}([s_i])$

We create a span representation for each constituent $c_k \in C(T_i)$. For the $k$th constituent, using its span indices $(i', j')$, we select its corresponding token representations $[\mathbf{h}^{\upsent}_{ii'}, \cdots, \mathbf{h}^{\upsent}_{ij'}]_k \in \mathbb{R}^{(j' - i') \times d}$. We then use span attention \cite{lee-etal-2017-end} to reduce this span to a fixed-length vector $\mathbf{h}^{\upspan}_{k}$. Finally, we compute deletion probabilities using a weight vector $\mathbf{w} \in \mathbb{R}^{d}$ as follows: $P(C_k^X = 1|s_j) = \sigma(\mathbf{h}^{\upspan\top}_{k} \mathbf{w})$, where $C_k^X$ is either a plausibility or salience random variable.

\subsection{Postprocessing}
As alluded to in Section~\ref{sec:syntactic-compression-rules}, there are certain cases where the syntactic compression rules license deleting a chain of constituents rather than individual ones. A common example of this is in conjoined noun phrases (NP$_1$--[CC--NP$_2$]) where if the second noun phrase NP$_2$ is deleted, its preceding coordinating conjunction CC can also be deleted without affecting the grammaticality of the sentence. To avoid changing the compression model substantially, we relegate secondary deletions to a postprocessing step, where if a primary constituent like NP$_2$ is deleted at test-time, its secondary constituents are also automatically deleted.

\section{Training and Inference}

The extraction and compression models in our summarization system are trained separately, but both used in a pipeline during inference. Because the summarization datasets we use do not come with labels for extraction and compression, we chiefly rely on structured oracles that provide supervision for our models. In this section, we describe our oracle design decisions, learning objectives, and inference procedures.\footnote{See Appendices \ref{sec:appendix-training-details} and \ref{sec:appendix-inference-details} for training and inference hyperparameters, respectively.}

\subsection{Extraction Supervision}
\label{sec:extraction-supervision}

Following \citet{liu-lapata-2019-text}, we derive an oracle extractive summary using a greedy algorithm that selects up to $k$ sentences in a document that maximize ROUGE \cite{lin-2004-rouge} with respect to the reference summary.\footnote{We found that using beam search to derive the oracle yielded higher oracle ROUGE, but also a significantly harder learning problem, and the extractive model trained on this oracle actually performed worse at test time.}

\subsection{Compression Supervision}
\label{sec:compression-supervision}

Because plausibility and salience are two different views of compression, as introduced in Section~\ref{sec:syntactic-compression-rules}, we have different methods for deriving their supervision. However, their oracles share the same high-level structure, which procedurally operate as follows: an oracle takes in as input an uncompressed sentence $x$, compressed sentence or paragraph $y$, and a similarity function $f$. Using the list of available compression rules $R(T_x)$ for $x$, if $x$ without a constituent $c_k \in R(T_x)$ results in $f(x \backslash c_k, y) > f(x, y)$, we assign $c_k$ a positive ``delete'' label, otherwise we assign it a negative ``keep'' label. Intuitively, this oracle measures whether the deletion of a constituent causes $x$ to become closer to $y$. We set $f$ to ROUGE \cite{lin-2004-rouge}, primarily for computational efficiency, although more complex similarity functions such as BERTScore \cite{zhang-2020-bertscore} could be used without modifying our core approach. Below, we elaborate on the nature of $x$ and $y$ for plausibility and salience, respectively.

\paragraph{Plausibility.} We leverage labeled, parallel sentence compression data from news headlines to learn plausibility. \citet{filippova-2013-overcoming} create a dataset of 200,000 news headlines and the lead sentence of its corresponding article, where each headline $x$ is a compressed extract of the lead sentence $y$. Critically, the headline is a subtree of the dependency relations induced by the lead sentence, ensuring that $x$ and $y$ will have very similar syntactic structure. \citet{filippova-2013-overcoming} further conduct a human evaluation of the headline and lead sentence pairs and conclude that, with 95\% confidence, annotators find the pairs ``indistinguishable'' in terms of readability and informativeness. This dataset therefore suits our purposes for plausibility as we have defined it.

\paragraph{Salience.} Though the sentence compression data described above offers a reasonable prior on span-level deletions, the salience of a particular deletion is a domain-dependent notion that should be learned from in-domain data. One way to approximate this is to consider whether the deletion of a span in a sentence $x_i$ of an extractive summary increases ROUGE with the reference summary $y$ \cite{xu-durrett-2019-neural}, allowing us to estimate what types of spans are \textit{likely} or \textit{unlikely} to appear in a summary. We can therefore derive salience labels directly from labeled summarization data.

\subsection{Learning}

In aggregate, our system requires training three models: an extraction model ($\theta_\textrm{E}$), a plausibility model ($\theta_\textrm{P}$), and a salience model ($\theta_\textrm{S}$).

The extraction model optimizes log likelihood over each selection decision $S_j$ in document $D_i$, defined as $\mathcal{L}_\textrm{EXT} = -\sum^{n}_{i=1} \sum_{j \in D_i} \log P(S_j^{(i)}=S_{j}^{(i)*}|D_i)$ where $S^{(i)*}_{j}$ is the gold label for selecting the $j$th sentence in the $i$th document.

The plausibility model optimizes log likelihood over the oracle decision $C^{\text{PL}(i)*}_{jk}$ for each constituent $c_k \in R(T_j)$ in sentence $j$, defined as $\mathcal{L}_\textrm{CMP} = -\sum^{m}_{j=1}\sum_{c_k \in R(T_j)} \log P(C^{\text{PL}(i)}_{jk}=C^{\text{PL}(i)*}_{jk}|s_j^{(i)})$. The salience model operates analogously over the $C^{\text{SAL}}$ variables.

\subsection{Inference}
\label{sec:inference}

While our sentence selection and compression stages are modeled independently, structurally we need to combine these decisions to yield a coherent summary, recognizing that these models have not been optimized directly for ROUGE.

Our pipeline consists of three steps: (1) For an input document $D$, we select the top-$k$ sentences with the highest posterior selection probabilities: $\mathrm{argmax}_k P(S_i=1 | D;\theta_\textrm{E})$. (2) Next, for each selected sentence $j$, we obtain plausible compressions $Z_\textrm{P} = \{c_k | P(C^{\text{PL}}_{jk} =1 | s_j;\theta_\textrm{P}) > \lambda_\textrm{P} , \forall c_k \in R(T_j) \}$ and salient compressions $Z_\textrm{S} = \{c_k | P(C^{\text{SAL}}_{jk} =1| s_j;\theta_\textrm{S}) > \lambda_\textrm{S} , \forall c_k \in R(T_j) \}$, where $\lambda_\textrm{P}$ and $\lambda_\textrm{S}$ are hyperparameters discovered with held-out samples. (3) Finally, we only delete constituent spans licensed by both the plausibility and salience models, denoted as $Z_\textrm{P} \cap Z_\textrm{S}$, for each sentence. The remaining tokens among all selected sentences form the compressive summary.\footnote{Our pipeline overall requires 3x more parameters than a standard Transformer-based extractive model (e.g., BERTSum). However, our compression module (which accounts for 2/3 of these parameters) can be applied on top of any off-the-shelf extractive model, so stronger extractive models with more parameters can be combined with our approach as well.}

We do not perform joint inference over the plausibility and salience models because plausibility is a necessary precondition in span-based deletion, as defined in Section~\ref{sec:plausibility}. If, for example, a compression has a low plausibility score but high salience score, it will get deleted during joint inference, but this may negatively affect the well-formedness of the summary. As we demonstrate in Section~\ref{sec:compression-analysis}, the plausibility model enforces strong guardrails that prevent the salience model from deleting arbitrary spans that result in higher ROUGE but at the expense of syntactic or semantic errors.

\begin{table*}[t]
\setlength{\tabcolsep}{4.85pt}
\small
\centering
\begin{tabular}{clrrrrrrrrrrrr}
\toprule
& & \multicolumn{3}{c}{CNN} & \multicolumn{3}{c}{WikiHow} & \multicolumn{3}{c}{XSum} & \multicolumn{3}{c}{Reddit} \\
 \cmidrule(lr){3-5} \cmidrule(lr){6-8} \cmidrule(lr){9-11} \cmidrule(lr){12-14}
Type & Model & R1 & R2 & RL & R1 & R2 & RL & R1 & R2 & RL & R1 & R2 & RL \\
\midrule
\modelext & Lead-$k$ & 29.80 & 11.40 & 26.45 & 24.96 & 5.83 & 23.23 & 17.02 & 2.72 & 13.79 & 19.64 & 2.40 & 14.79 \\
\modelext & BERTSum & --- & --- & --- & 30.31 & 8.71 & 28.24 & 22.86 & 4.48 & 17.16 & 23.86 & 5.85 & 19.11 \\
\modelext & MatchSum$^\diamondsuit$ & --- & --- & --- & 31.85 & 8.98 & 29.58 & 24.86 & 4.66 & 18.41 & 25.09 & 6.17 & 20.13 \\
\modelabs & PEGASUS$_\textrm{BASE}$ & --- & --- & --- & 36.58 & 15.64 & 30.01 & 39.79 & 16.58 & 31.70 & 24.36 & 6.09 & 18.75 \\ 
\modelabs & PEGASUS$^\heartsuit_\textrm{LARGE}$ & --- & --- & --- & \textbf{43.06} & \textbf{19.71} & \textbf{34.80} & \textbf{47.21} & \textbf{24.56} & \textbf{39.25} & \textbf{26.63} & \textbf{9.01} & \textbf{21.60} \\
\midrule
\modelext & \ext & 33.12 & 13.88 & 29.51 & 30.94 & 9.06 & 28.81 & 24.23 & 4.95 & 18.30 & 24.42 & 6.10 & 19.57 \\
\modelcmp & CUPS & \textbf{35.22} & \textbf{14.19} & \textbf{31.51} & \textbf{32.43} & \textbf{9.44} & \textbf{30.24} & \textbf{26.04} & \textbf{5.36} & \textbf{19.90} & \textbf{25.99} & \textbf{6.57} & \textbf{21.08} \\
\bottomrule
\end{tabular}
\caption{\textbf{Results on CNN, WikiHow, XSum, and Reddit.} Our system consistently achieves higher ROUGE than extraction-only baselines. Additionally, our system achieves higher ROUGE-L than PEGASUS$_\textrm{BASE}$ on WikiHow and Reddit without summarization-specific pre-training. $^\diamondsuit$Extractive SOTA; $^\heartsuit$Abstractive SOTA.}
\label{tab:other-id-results}
\end{table*}

\begin{table}[t]
\setlength{\tabcolsep}{5pt}
\small
\centering
\begin{tabular}{clrrr}
\toprule
Type & Model & R1 & R2 & RL \\
\midrule
\modelext & Lead-3 & 40.42 & 17.62 & 36.67 \\
\modelext & BERTSum & 43.25 & 20.24 & 39.63 \\
\modelext & MatchSum$^\diamondsuit$ & \textbf{44.41} & 20.86 & 40.55 \\
\modelabs & PEGASUS$_\textrm{BASE}$ & 41.79 & 18.81 & 38.93 \\
\modelabs & PEGASUS$^\heartsuit_\textrm{LARGE}$ & 44.17 & \textbf{21.47} & \textbf{41.11} \\
\midrule
\modelext & \ext (BERT) & 43.16 & 20.10 & 39.52 \\
\modelext & \ext & 43.65 & 20.57 & 40.02 \\
\modelcmp & CUPS & 44.02 & 20.57 & 40.38 \\
\modelcmp & MatchSum + \cmp & \textbf{44.69} & \textbf{20.71} & \textbf{40.86} \\
\bottomrule
\end{tabular}
\caption{\textbf{Results on CNN/DM.} Notably, a pipeline with MatchSum \cite{zhong-etal-2020-matchsum} extraction and our compression module achieves state-of-the-art ROUGE-1. $^\diamondsuit$Extractive SOTA; $^\heartsuit$Abstractive SOTA.}
\label{tab:cnndm-id-results}
\end{table}

\section{Experimental Setup}

We benchmark our system first with an automatic evaluation based on ROUGE-1/2/L F$_1$ \cite{lin-2004-rouge}.\footnote{Following previous work, we use \texttt{pyrouge} with the default command-line arguments: \texttt{-c 95 -m -n 2}} Our experiments use the following English datasets: CNN/DailyMail \cite{hermann-2015-teaching}, CNN (subset of CNN/DM), New York Times \cite{sandhaus-2008-the}, WikiHow \cite{koupaee-wang-2018-wikihow}, XSum \cite{narayan-2018-xsum}, and Reddit \cite{kim-2019-reddit}.\footnote{See Appendix~\ref{sec:appendix-summarization-datasets} for dataset splits.}

We seek to answer three questions: (1) How does our compressive system stack up against our own extractive baseline and past extractive approaches? (2) Do our plausibility and salience modules successfully model their respective phenomena? (3) How can these pieces be used to improve cross-domain summarization?

\paragraph{Systems for Comparison.} We refer to our full compressive system as \textbf{CUPS}\footnote{\textbf{C}ompressive S\textbf{u}mmarization with \textbf{P}lausibility and \textbf{S}alience}, which includes \ext and \cmp, the extraction and compression components, respectively. \ext is a re-implementation of BERTSum \cite{liu-2019-single} and \cmp is a module consisting of both the plausibility and salience models. The pre-trained encoders in the extraction and compression modules are set to ELECTRA$_\textrm{BASE}$ \cite{clark-2020-electra}, unless specified otherwise.

Because our approach is fundamentally extractive (albeit with compression), we chiefly compare against state-of-the-art extractive models: \textbf{BERTSum} \cite{liu-2019-single}, the canonical architecture for sentence-level extraction with pre-trained encoders, and \textbf{MatchSum} \cite{zhong-etal-2020-matchsum}, a summary-level semantic matching model that uses BERTSum to prune irrelevant sentences. These models outperform recent compressive systems \cite{xu-durrett-2019-neural,mendes-etal-2019-jointly}; updating the architectures of these models and extending their oracle extraction procedures to the range of datasets we consider is not straightforward.

To contextualize our results, we also compare against a state-of-the-art abstractive model, \textbf{PEGASUS} \cite{zhang-2020-pegasus}, a seq2seq Transformer pre-trained with ``gap-sentences.'' This comparison is not entirely apples-to-apples, as this pre-training objective uses very large text corpora (up to 3.8TB) in a summarization-specific fashion. We expect our approach to stack with further advances in pre-training. 

Extractive, abstractive, and compressive approaches are typed as \modelext, \modelabs, and \modelcmp, respectively, throughout the experiments.

\section{In-Domain Experiments}

\begin{table*}
\small
\centering
\begin{tabular}{p{15.4cm}l}
\toprule
opening statements in the murder trial of movie theater massacre suspect james holmes are scheduled for april 27\compress{, more than a month ahead of schedule}, a colorado court spokesman said. holmes\compress{, 27,} is charged as the \compress{sole} gunman who stormed a crowded movie theater \compress{at a midnight showing of "the dark knight rises" in aurora, colorado}, and opened fire\compress{, killing 12 people and wounding 58 more in july 2012}. holmes\compress{, a one-time neuroscience doctoral student,} faces 166 counts\compress{, including murder and attempted murder charges}. \\
\midrule
the accident happened in santa ynez  california\compress{, near where crosby lives}. crosby was driving at \compress{approximately} 50 mph when he struck the jogger, according to california highway patrol spokesman don clotworthy. the jogger suffered multiple fractures, and was airlifted to a hospital \compress{in santa barbara}, clotworthy said. \\
\midrule
update: jonathan hyla said \compress{in an phone interview} monday that his interview with cate blanchett was mischaracterized \compress{when an edited version went viral around the web last week}. ``she wasn't upset,'' he told cnn. blanchett ended the interview laughing\compress{, hyla said,} and ``she was in on the joke.'' \\
\bottomrule
\end{tabular}
\caption{CUPS-produced summaries on CNN, where \compress{strikethrough text} implies the span is deleted as judged by the plausibility and salience models. The base sentences before applying compression are derived from \ext, the sentence extractive model.}
\label{tab:cnn-samples}
\end{table*}

\subsection{Benchmark Results}

Table~\ref{tab:other-id-results} (CNN, WikiHow, XSum, Reddit) and \ref{tab:cnndm-id-results} (CNN/DM) show ROUGE results. From these tables, we make the following observations:

\paragraph{Compression consistently improves ROUGE, even when coupled with a strong extractive model.} Across the board, we see improvements in ROUGE when using CUPS. Our results particularly contrast with recent trends in compressive summarization where span-based compression (in joint and pipelined forms) \textit{decreases} ROUGE over sentence extractive baselines \cite{zhang-etal-2018-latsum,mendes-etal-2019-jointly}. Gains are especially pronounced on datasets with more abstractive summaries, where applying compression roughly adds +2 ROUGE-1; however, we note there is a large gap between extractive and abstractive approaches on tasks like XSum due to the amount of paraphrasing in reference summaries \cite{narayan-2018-xsum}. Nonetheless, our system outperforms strong extractive models on these datasets, and also yields competitive results on CNN/DM. In addition, Table~\ref{tab:cnn-samples} includes representative summaries produced by our compressive system. The summaries are highly compressive: spans not contributing to the main event or story are deleted, while maintaining grammaticality and factuality.

\paragraph{Our compression module can also improve over other off-the-shelf extractive models.} The pipelined nature of our approach allows us to replace the current BERTSum \cite{liu-lapata-2019-text} extractor with any arbitrary, black-box model that retrieves important sentences. We apply our compression module on system outputs from MatchSum \cite{zhong-etal-2020-matchsum}, the current state-of-the-art extractive model, and also see gains in this setting with no additional modification to the system.

\subsection{Plausibility Study}
\label{sec:plausibility-study}

Given that our system achieves high ROUGE, we now investigate whether its compressed sentences are grammatical and factual. The plausibility model is responsible for modeling these phenomena, as defined in Section~\ref{sec:plausibility}, thus we analyze its compression decisions in detail. Specifically, we run the plausibility model on 50 summaries from each of CNN and Reddit, and have annotators judge whether the predicted plausible compressions are grammatical and factual with respect to the original sentence.\footnote{See Appendix~\ref{sec:appendix-plausibility-study} for further information on the annotation task and agreement scores.} By nature, this evaluates the \textit{precision} of span-based deletions.

Because the plausibility model uses candidate spans from the high-recall compression rules (defined in Section~\ref{sec:syntactic-compression-rules}), we compare our plausibility model against the baseline consisting of simply the spans identified by these rules. The results are shown in Table~\ref{tab:human-eval}. On both CNN and Reddit, the plausibility model's deletions are highly grammatical, and we also see evidence that the plausibility model makes more semantically-informed deletions to maintain factuality, especially on CNN. 

\begin{table}[t]
\setlength{\tabcolsep}{4pt}
\small
\centering
\begin{tabular}{lrrrr}
\toprule
 & \multicolumn{2}{c}{CNN} & \multicolumn{2}{c}{Reddit} \\
\cmidrule(lr){2-3} \cmidrule(lr){4-5}
System & G & F & G & F \\
\midrule
Compression Rules & 87.9 & 75.7 & 73.5 & 60.8 \\
\quad + Plausibility Model & \textbf{96.0} & \textbf{89.7} & \textbf{93.1} & \textbf{66.7} \\
\bottomrule
\end{tabular}
\caption{Human evaluation of grammaticality (G) and factuality (F) of summaries, comparing the precision of span deletions from our compression rules (\S\ref{sec:syntactic-compression-rules}) before and after applying the plausibility model (\S\ref{sec:plausibility}).}
\label{tab:human-eval}
\end{table}

Factuality performance is lower on Reddit, but incorporating the plausibility model on top of the compression rules results in a 6\% gain in precision. There is still, however, a large gap between factuality in this setting and factuality on CNN, which we suspect is because Reddit summaries are different in style and structure than CNN summaries: they largely consist of short event narratives \cite{kim-2019-reddit}, and so annotators may disagree on the degree to which deleting spans such as subordinate clauses impact the meaning of the events described.

\begin{table*}[t]
\setlength{\tabcolsep}{4pt}
\small
\centering
\begin{tabular}{clrrrrrrrrrrrrrrr}
\toprule
& & \multicolumn{3}{c}{NYT $\rightarrow$ CNN} & \multicolumn{3}{c}{CNN $\rightarrow$ Reddit} & \multicolumn{3}{c}{XSum $\rightarrow$ WikiHow} & \multicolumn{3}{c}{Average} \\
\cmidrule(lr){3-5} \cmidrule(lr){6-8} \cmidrule(lr){9-11} \cmidrule(lr){12-14}
Type & Model & R1 & R2 & RL & R1 & R2 & RL & R1 & R2 & RL & R1 & R2 & RL \\
\midrule
\multicolumn{10}{l}{In-Domain}  \\
\midrule
\modelext & Lead-$k$ & 29.80 & 11.40 & 26.45 & 19.64 & 2.40 & 14.79 & 24.96 & 5.83 & 23.23 & 24.80 & 6.54 & 21.49 \\
\modelext & \ext & \textbf{33.12} & \textbf{13.88} & \textbf{29.51} & \textbf{24.42} & \textbf{6.10} & \textbf{19.57} & \textbf{30.94} & \textbf{9.06} & \textbf{28.81} & \textbf{29.49} & \textbf{9.68} & \textbf{25.96} \\
\midrule
\multicolumn{10}{l}{Out-of-Domain} \\
\midrule
\modelext & \ext & 31.05 & 12.46 & 27.64 & 21.32 & 4.54 & 17.08 & 28.32 & 7.54 & 26.35 & 26.90 & 12.27 & 23.69 \\
& \quad + Fine-Tune (500) & 31.90 & 13.04 & 28.42 & 23.76 & 5.66 & 18.95 & 29.44 & 8.25 & 27.41 & 28.37 & 8.98 & 24.93 \\
\modelcmp & CUPS & 31.98 & 12.77 & 28.53 & 22.25 & 4.82 & 17.94 & 29.17 & 7.65 & 27.28 & 27.80 & 8.41 & 24.59 \\
& \quad + Fine-Tune (500) & \textbf{33.98} & \textbf{13.25} & \textbf{30.39} & \textbf{25.01} & \textbf{5.96} & \textbf{20.10} & \textbf{30.52} & \textbf{8.44} & \textbf{28.48} & \textbf{29.84} & \textbf{9.22} & \textbf{26.32} \\
\bottomrule
\end{tabular}
\caption{\textbf{Results on out-of-domain transfer tasks.} Fine-tuning results are averaged across 5 runs, each with a random batch of 500 target domain samples. Variance among these runs is very low; see Appendix~\ref{sec:appendix-ood-results}.}
\label{tab:od-results}
\end{table*}

\begin{figure}[t]
    \centering
    \includegraphics[width=6cm]{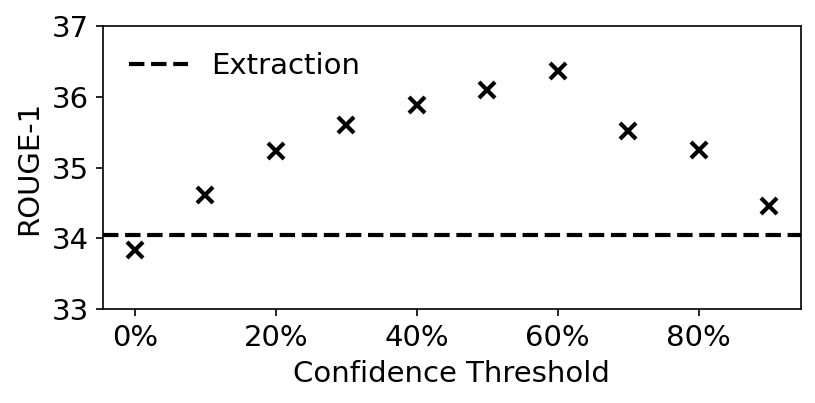}
    \caption{Varying the salience threshold $\lambda_\textrm{S} \in [0, 1)$ (depicted as \% confidence) and its impact on ROUGE upon deleting spans $Z_\textrm{P} \cap Z_\textrm{S}$.}
    \label{fig:salience-confidence}
\end{figure}

\subsection{Compression Analysis}
\label{sec:compression-analysis}

The experiments above demonstrate the plausibility model generally selects spans that, if deleted, preserve grammaticality and factuality. In this section, we dive deeper into how the plausibility and salience models work together in the final trained summary model, presenting evidence of typical compression patterns. We analyze (1) our default system CUPS, which deletes spans $Z_\textrm{P} \cap Z_\textrm{S}$; and (2) a variant CUPS-\textsc{NoPl} (\textit{without} plausibility but \textit{with} salience), which only deletes spans $Z_\textrm{S}$, to specifically understand what compressions the salience model makes without the plausibility model's guardrails. Using 100 randomly sampled documents from CNN, we conduct a series of experiments detailed below.



\paragraph{On average, per sentence, 16\% of candidate spans deleted by the salience model alone are not plausible.} For each sentence, our system exposes a list of spans for deletion, denoted by $Z_\textrm{P} \cap Z_\textrm{S}$ and $Z_\textrm{S}$ for CUPS and CUPS-\textsc{NoPl}, respectively. Because $Z_\textrm{S}$ is identical across both variants, we can compute the plausibility model's \textit{rejection rate} (16\%), defined as $|Z_\textrm{S} \cap Z_\textrm{P}^C| / |Z_\textrm{S}|$. Put another way, how many compressions does the plausibility model \textit{reject} if partnered with the salience model? On average, per sentence, the plausibility model rejects 16\% of spans approved by the salience model alone, so it does non-trivial filtering of the compressions. We observe a drop in the token-level compression ratio, from 26\% in CUPS to 24\% in CUPS-\textsc{NoPl}, which is partially a result of this. From a ROUGE-1/2 standpoint, the slight reduction in compression yields a peculiar effect: on this subset of summaries, CUPS achieves 36.23/14.61 while CUPS-\textsc{NoPl} achieves 36.1/14.79, demonstrating the plausibility model trades off some salient deletions (-R1) for overall grammaticality (+R2) \cite{paulus-2018-a}.




\paragraph{Using salience to discriminate between plausible spans increases ROUGE.} With CUPS, we perform a line search on $\lambda_\textrm{S} \in [0, 1)$, which controls the confidence threshold for deleting non-salient spans as described in Section~\ref{sec:inference}.\footnote{Our assumption is that posterior probabilities are calibrated, which holds true for various pre-trained Transformers across a range of tasks \cite{desai-durrett-2020-calibration}.} Figure~\ref{fig:salience-confidence} shows ROUGE-1 across multiple salience cutoffs. When $\lambda_\textrm{S}=0$, all plausible spans are deleted; in terms of ROUGE, this setting underperforms the extractive baseline, indicating we end up deleting spans that contain pertinent information. In contrast, at the peak when $\lambda_\textrm{S}=0.6$, we delete non-salient spans with at least 60\% confidence, and obtain considerably better ROUGE. These results indicate that the spans selected by the plausibility model are fundamentally good, but the ability to weigh the content relevance of these spans is critical to end-task performance.

\section{Out-of-Domain Experiments}

Additionally, we examine the cross-domain generalizability of our compressive summarization system. We set up three source $\rightarrow$ target transfer tasks guided by real-world settings: (1) NYT $\rightarrow$ CNN (one newswire outlet to another), (2) CNN $\rightarrow$ Reddit (newswire to social media, a low-resource domain), and (3) XSum $\rightarrow$ WikiHow (single to multiple sentence summaries with heavy paraphrasing).

For each transfer task, we experiment with two types of settings: (1) \textbf{zero-shot} transfer, where our system with parameters $[\theta_\textrm{E};\theta_\textrm{P};\theta_\textrm{S}]$ is directly evaluated on the target test set; and (2) \textbf{fine-tuned} transfer, where $[\theta_\textrm{E};\theta_\textrm{S}]$ are fine-tuned with 500 target samples, then the resulting system with parameters $[\theta'_\textrm{E};\theta_\textrm{P};\theta'_\textrm{S}]$ is evaluated on the target test set. As defined in Section~\ref{sec:plausibility}, plausibility is a domain-independent notion, thus we do not fine-tune $\theta_\textrm{P}$.

Table~\ref{tab:od-results} shows the results. Our system maintains strong zero-shot out-of-domain performance despite distribution shifts: extraction outperforms the lead-$k$ baseline, and compression adds roughly +1 ROUGE-1. This increase is largely due to compression improving ROUGE precision: extraction is adept at retrieving content-heavy sentences with high recall, and compression helps focus on salient content within those sentences.

More importantly, we see that \textbf{performance via fine-tuning on 500 samples matches or exceeds in-domain extraction ROUGE.} On NYT $\rightarrow$ CNN and CNN $\rightarrow$ Reddit, our system outperforms in-domain extraction baselines (trained on tens of thousands of examples), and on XSum $\rightarrow$ WikiHow, it comes within 0.3 in-domain average ROUGE. These results suggest that our system could be applied widely by crowdsourcing a relatively small number of summaries in a new domain.

\section{Related Work}

\paragraph{Compressive Summarization.} Our work follows in a line of systems that use auxiliary training data or objectives to learn sentence compression \cite{martins-smith-2009-summarization,woodsend-lapata-2012-multi-aspect,qian-liu-2013-fast}. Unlike these past approaches, our compression system uses \textit{both} a plausibility model optimized for grammaticality and a salience model optimized for ROUGE. \citet{almeida-martins-2013-fast} leverage such modules and learn them jointly in a multi-task learning setup, but face an intractable inference problem in their model which needs sophisticated approximations. Our approach, by contrast, does not need such approximations or expensive inference machinery like ILP solvers \cite{martins-smith-2009-summarization,berg-kirkpatrick-2011-jointly,durrett-etal-2016-learning}. The highly decoupled nature of our pipelined compressive system is an advantage in terms of training simplicity: we use only simple MLE-based objectives for extraction and compression, as opposed to recent compressive methods that use joint training \cite{xu-durrett-2019-neural,mendes-etal-2019-jointly} or reinforcement learning \cite{zhang-etal-2018-latsum}. Moreover, we demonstrate our compression module can stack with state-of-the-art sentence extraction models, achieving additional gains in ROUGE.

One significant line of prior work in compressive summarization relies on heavily engineered rules for syntactic compression \cite{berg-kirkpatrick-2011-jointly,li-etal-2014-improving-multi,wang-etal-2013-a,xu-durrett-2019-neural}. By relying on our data-driven objectives to ultimately perform compression, our approach can rely on a leaner, much more minimal set of constituency rules to extract candidate spans.

\citet{gehrmann-2018-bottomup} also extract sub-sentential spans in a ``bottom-up'' fashion, but their method does not incorporate grammaticality and only works best with an abstractive model; thus, we do not compare to it in this work.

\paragraph{Discourse-based Compression.} Recent work also demonstrates elementary discourse units (EDUs), spans of sub-sentential clauses, capture salient content more effectively than entire sentences \cite{hirao-etal-2013-knapsack,li-etal-2016-edu,durrett-etal-2016-learning,xu-etal-2020-discobert}. Our approach is significantly more flexible because it does not rely on an a priori chunking of a sentence, but instead can delete variably sized spans based on what is contextually permissible. Furthermore, these approaches require RST discourse parsers and in some cases coreference systems \cite{xu-etal-2020-discobert}, which are less accurate than the constituency parsers we use.

\section{Conclusion}

In this work, we present a compressive summarization system that decomposes span-level compression into two learnable objectives, plausibility and salience, on top of a minimal set of rules derived from a constituency tree. Experiments across both in-domain and out-of-domain settings demonstrate our approach outperforms strong extractive baselines while creating well-formed summaries.

\section*{Acknowledgments}

This work was partially supported by NSF Grant IIS-1814522, NSF Grant SHF-1762299, a gift from Salesforce Inc., and an equipment grant from NVIDIA. The authors acknowledge the Texas Advanced Computing Center (TACC) at The University of Texas at Austin for providing HPC resources used to conduct this research. Results presented in this paper were obtained using the Chameleon testbed supported by the National Science Foundation. Thanks as well to the anonymous reviewers for their helpful comments.

\bibliography{emnlp2020}
\bibliographystyle{acl_natbib}

\newpage
\appendix

\setcounter{table}{0}
\setcounter{figure}{0}

\section{Summarization Datasets}
\label{sec:appendix-summarization-datasets}

Table~\ref{tab:appendix-dataset-splits} lists training, development, and test splits for each dataset used in our experiments.

\begin{table}[h]
\setlength{\tabcolsep}{4pt}
\centering
\small
\begin{tabular}{lrrrrr}
\toprule
Dataset & $k$ & Train & Dev & Test \\
\midrule
CNN/Daily Mail & 3 & 287,084 & 13,367 & 11,489 \\
\quad CNN & 3 & 90,266 & 1,220 & 1,093 \\
New York Times & 3 & 137,772 & 17,222 & 17,220 \\
XSum & 2 & 203,028 & 11,273 & 11,332 \\
WikiHow & 4 & 168,126 & 6,000 & 6,000 \\
Reddit & 2 & 41,675 & 645 & 645 \\
\bottomrule
\end{tabular}
\caption{Training, development, and test dataset sizes for CNN/Daily Mail \cite{hermann-2015-teaching}, CNN (subset of CNN/DM), New York Times \cite{sandhaus-2008-the}, XSum \cite{narayan-2018-xsum}, WikiHow \cite{koupaee-wang-2018-wikihow}, and Reddit \cite{kim-2019-reddit}. For each dataset, the extraction model selects the top-$k$ sentences to form the basis of the compressive summary.}
\label{tab:appendix-dataset-splits}
\end{table}

\begin{table}[t]
\setlength{\tabcolsep}{4pt}
\centering
\small
\begin{tabular}{lrr}
\toprule
Hyperparameter & Extraction & Compression \\
\midrule
Train Steps & 10,000 & 10,000 \\
Eval Steps & 1,000 & 1,000 \\
Eval Interval & 1,000 & 1,000 \\
Batch Size & 16 & 16 \\
Learning Rate & 1e-5 & 1e-5 \\
Optimizer & AdamW & AdamW \\
Weight Decay & 0 & 0 \\
Gradient Clip & 1.0 & 1.0 \\
Max Sequence Length & 512 & 256 \\
Max Spans & --- & 50 \\
\bottomrule
\end{tabular}
\caption{Training hyperparameters for the extraction and compression models (\S\ref{sec:summarization-system}).}
\label{tab:appendix-training-hyperparams}
\end{table}

\begin{table}[t]
\setlength{\tabcolsep}{3.5pt}
\centering
\small
\begin{tabular}{lrrrrr}
\toprule
Encoder & CNN/DM & CNN & WikiHow & XSum & Reddit \\
\midrule
\multicolumn{6}{l}{Hyperparameter: Plausibility ($\lambda_\textrm{P}$)} \\
\midrule
BERT & 0.6 & 0.6 & 0.6 & 0.6 & 0.6 \\
ELECTRA & 0.6 & 0.6 & 0.6 & 0.6 & 0.6 \\
\midrule
\multicolumn{6}{l}{Hyperparameter: Salience ($\lambda_\textrm{S}$)} \\
\midrule
BERT & 0.7 & 0.5 & 0.4 & 0.55 & 0.65 \\
ELECTRA & 0.7 & 0.5 & 0.45 & 0.6 & 0.7 \\
\bottomrule
\end{tabular}
\caption{BERT- and ELECTRA-based system hyperparameters for the plausibility (\S\ref{sec:plausibility}) and salience models (\S\ref{sec:salience}). We fix the plausibility threshold at 0.6 and only optimize the salience thresold.}
\label{tab:appendix-inf-cmp-hyperparams}
\end{table}

\section{Training Details}
\label{sec:appendix-training-details}

Table~\ref{tab:appendix-training-hyperparams} details the hyperparameters for training the extraction and compression models. These hyperparameters largely borrowed from previous work \cite{devlin-etal-2019-bert}, and we do not perform any additional grid searches in the interest of simplicity. The pre-trained encoders are set to either \texttt{\small{bert-base-uncased}} or \texttt{\small{google/electra-base-discriminator}} from HuggingFace Transformers \cite{wolf2019huggingfaces}. Following previous work \cite{liu-2019-single,zhong-etal-2020-matchsum}, we use the best performing model among the top three validation checkpoints.

\section{Inference Details}
\label{sec:appendix-inference-details}

Our system uses two hyperparameters at test-time to control the level of compression performed by the plausibility and salience models. Table \ref{tab:appendix-inf-cmp-hyperparams} shows the BERT- and ELECTRA-based system hyperparameters, respectively. We sweep the salience model threshold $\lambda_\textrm{S} \in [0.1, 0.9]$ with a granularity of 0.05; across all datasets used in the in-domain experiments (CNN/DM, CNN, WikiHow, XSum, and Reddit), this process takes roughly 8 hours on a 32GB NVIDIA V100 GPU. 

\section{Plausibility Study}
\label{sec:appendix-plausibility-study}

\begin{table}[h]
\setlength{\tabcolsep}{4pt}
\small
\centering
\begin{tabular}{lrr}
\toprule
Study & CNN & Reddit \\
 \midrule
Grammaticality & 0.24 & 0.17 \\
Factuality & 0.28 & 0.34 \\
\bottomrule
\end{tabular}
\caption{Annotator agreement for grammaticality and factuality studies on CNN and Reddit. Values displayed are computed using Krippendorff's $\alpha$ \cite{krippendorff-1980-alpha}.}
\label{tab:appendix-agreement}
\end{table}

\begin{table*}[t]
\setlength{\tabcolsep}{3pt}
\small
\centering
\begin{tabular}{clrrrrrrrrrrrrrrr}
\toprule
& & \multicolumn{3}{c}{CNN/DM} & \multicolumn{3}{c}{CNN} & \multicolumn{3}{c}{WikiHow} & \multicolumn{3}{c}{XSum} & \multicolumn{3}{c}{Reddit} \\
\cmidrule(lr){3-5} \cmidrule(lr){6-8} \cmidrule(lr){9-11} \cmidrule(lr){12-14} \cmidrule(lr){15-17}
Type & Model & R1 & R2 & RL & R1 & R2 & RL & R1 & R2 & RL & R1 & R2 & RL & R1 & R2 & RL \\
\midrule
\modelext & \ext & 43.16 & 20.10 & 39.52 & 32.41 & 13.59 & 28.93 & 30.45 & 8.74 & 28.34 & 23.59 & 4.55 & 17.81 & 23.87 & 5.84 & 19.27 \\
\modelcmp & CUPS & \textbf{43.55} & \textbf{20.11} & \textbf{39.93}  & \textbf{34.54} & \textbf{13.67} & \textbf{31.00} & \textbf{31.98} & \textbf{8.95} & \textbf{29.88} & \textbf{25.59} & \textbf{4.93} & \textbf{19.67} & \textbf{25.24} & \textbf{6.12} & \textbf{20.60} \\
\bottomrule
\end{tabular}
\caption{Results on CNN/DM, CNN, WikiHow, XSum, and Reddit with initializing the pre-trained encoders in CUPS to BERT$_\textrm{BASE}$ as opposed to ELECTRA$_\textrm{BASE}$.}
\label{tab:appendix-bert-results}
\end{table*}

\begin{table*}[t]
\setlength{\tabcolsep}{4pt}
\small
\centering
\begin{tabular}{clrrrrrrrrr}
\toprule
& & \multicolumn{3}{c}{WikiHow} & \multicolumn{3}{c}{XSum} & \multicolumn{3}{c}{Reddit} \\
 \cmidrule(lr){3-5} \cmidrule(lr){6-8} \cmidrule(lr){9-11}
Type & Model & R1 & R2 & RL & R1 & R2 & RL & R1 & R2 & RL \\
\midrule
\modelcmp & CUPS & 32.43 & \textbf{9.44} & 30.24 & 26.04 & \textbf{5.36} & \textbf{19.90} & 25.99 & 6.57 & 21.08 \\
\modelcmp & MatchSum + \cmp & \textbf{32.83} & 9.24 & \textbf{30.53} & \textbf{26.42} & 5.09 & 19.76 & \textbf{26.60} & \textbf{6.60} & \textbf{21.43} \\
\bottomrule
\end{tabular}
\caption{Results on WikiHow, XSum, and Reddit with replacing \ext with MatchSum \cite{zhong-etal-2020-matchsum}, a state-of-the-art extractive model.}
\label{tab:appendix-matchsum}
\end{table*}

\begin{table*}[t]
\setlength{\tabcolsep}{4pt}
\small
\centering
\begin{tabular}{clrrrrrrrrrrrr}
\toprule
 &  & \multicolumn{3}{c}{CNN} & \multicolumn{3}{c}{WikiHow} & \multicolumn{3}{c}{XSum} & \multicolumn{3}{c}{Reddit} \\
\cmidrule(lr){3-5} \cmidrule(lr){6-8} \cmidrule(lr){9-11} \cmidrule(lr){12-14}
Type & Model & R1 & R2 & RL & R1 & R2 & RL & R1 & R2 & RL & R1 & R2 & RL \\
\midrule
\modelcmp & CUPS & 35.22 & \textbf{14.19} & 31.51 & 32.43 & \textbf{9.44} & 30.24 & 26.04 & \textbf{5.36} & 19.90 & 25.99 & \textbf{6.57} & 21.08 \\
 & \quad - Plausibility & \textbf{35.29} & 14.03 & \textbf{31.63} & \textbf{32.54} & 9.34 & \textbf{30.36} & \textbf{26.36} & 5.35 & \textbf{20.19} & \textbf{26.11} & 6.56 & \textbf{21.19} \\
\bottomrule
\end{tabular}
\caption{Results on CNN, WikiHow, XSum, and Reddit with removing the plausibility model in \cmp.}
\label{tab:appendix-plausibility-ablation}
\end{table*}

We conduct our human evaluation on Amazon Mechanical Turk, and set up the following requirements: annotators must (1) reside in the US; (2) have a HIT acceptance rate $\ge$ 95\%; and (3) complete at least 50 HITs prior to this one. Each HIT comes with detailed instructions (including a set of representative examples) and 6 assignments. One of these assignments is a randomly chosen example from the instructions (the \textit{challenge} question), and the other five are samples we use in our actual study. In each assignment, annotators are presented with the original sentence and a candidate span, and asked if deleting the span negatively impacts the grammaticality and factuality of the resulting, compressed sentence. Each annotator is paid 50 cents upon completing the HIT; this pay rate was calibrated to pay roughly \$10/hour.

After all assignments are completed, we filter low-quality annotators according to two heuristics. An annotator is removed if he/she completes the assignment in under 60 seconds or answers the challenge question incorrectly. We see a substantial increase in agreement for both the grammaticality and factuality studies among the remaining annotators. The absolute agreement scores, as measured by Krippendorff's $\alpha$ \cite{krippendorff-1980-alpha}, are shown in Table~\ref{tab:appendix-agreement}. Consistent with prior grammaticality evaluations in summarization \cite{xu-durrett-2019-neural,xu-etal-2020-discobert}, agreement scores are objectively low due to the difficulty of the tasks, thus we compare the annotations with expert judgements. An expert annotator (one of the authors of this paper uninvolved with the development of the plausibility model) performed the CNN annotation task; we find, by using the majority vote among the crowdsourced annotations, the regular and expert annotators concur 80\% of the time on grammaticality and 60\% of the time on factuality; this establishes a higher degree of confidence in the crowdsourced annotations when aggregated.

\section{System Results with BERT}
\label{sec:appendix-bert-results}

Table~\ref{tab:appendix-bert-results} (CNN/DM, CNN, WikiHow, XSum, Reddit) shows results using BERT$_\textrm{BASE}$ as the pre-trained encoder. While the absolute ROUGE results with BERT$_\textrm{BASE}$ are lower than with ELECTRA$_\textrm{BASE}$, we still see a large improvement compared to the sentence extractive baseline.

\section{Extended MatchSum Results}
\label{sec:appendix-matchsum-results}

On WikiHow, XSum, and Reddit, we additionally experiment with replacing the sentences extracted from \ext with MatchSum \cite{zhong-etal-2020-matchsum} system outputs. From the results (see Table~\ref{tab:appendix-matchsum}), we see that our system with MatchSum extraction achieves the most gains on Reddit, but its average performance on WikiHow and XSum is more comparable to the standard CUPS system.

\section{Plausibility Ablation}

Table~\ref{tab:appendix-plausibility-ablation} shows results on CNN, WikiHow, XSum, and Reddit with removing the plausibility model in \cmp. Consistent with the analysis in Section~\ref{sec:compression-analysis}, we see the plausibility model is primarily responsible for gains in ROUGE-2, but in its absence, the salience model can delete arbitrary spans, resulting in gains in ROUGE-1 and ROUGE-L. This ablation demonstrates the need to analyze summaries outside of ROUGE since notions of grammaticality and factuality cannot easily be ascertained by computing lexical overlap with a reference summary.

\section{Out-of-Domain Results}
\label{sec:appendix-ood-results}

In Tables~\ref{tab:appendix-ood-nytcnn-var}, \ref{tab:appendix-ood-cnnreddit-var}, and \ref{tab:appendix-ood-xsumwikihow-var}, we show ROUGE results with standard deviations across 5 independent runs, for the fine-tuning experiments on NYT $\rightarrow$ CNN, CNN $\rightarrow$ Reddit, and XSum $\rightarrow$ WikiHow, respectively. Despite fine-tuning with a random batch of 500 samples each time, we consistently see low variance across the runs, demonstrating our system does not have an affinity towards particular samples in an out-of-domain setting.

Furthermore, we present an ablation of salience for the aforementioned transfer tasks in Table~\ref{tab:appendix-ood-ablations}. On NYT $\rightarrow$ CNN, salience only helps increase ROUGE-L, but we see consistent increases in average ROUGE on CNN $\rightarrow$ Reddit and XSum $\rightarrow$ WikiHow. We can expect larger gains by fine-tuning salience on more samples, but even with 500 out-of-domain samples, our compression module benefits from the inclusion of the salience model.

\begin{table*}[t]
\setlength{\tabcolsep}{4pt}
\small
\centering
\begin{tabular}{clrrr}
\toprule
& & \multicolumn{3}{c}{NYT $\rightarrow$ CNN} \\
\cmidrule(lr){3-5}
Type & Model & R1 (std) & R2 (std) & RL (std) \\
\midrule
\modelext & \ext & 33.74 (0.08) & 13.19 (0.11) & 30.46 (0.11) \\
\modelcmp & CUPS & 33.98 (0.06) & 13.25 (0.11) & 30.39 (0.07) \\
\bottomrule
\end{tabular}
\caption{Results on NYT $\rightarrow$ CNN, reporting ROUGE with standard deviation across 5 independent runs with a random batch of 500 samples.}
\label{tab:appendix-ood-nytcnn-var}
\end{table*}

\begin{table*}[t]
\setlength{\tabcolsep}{4pt}
\small
\centering
\begin{tabular}{clrrr}
\toprule
& & \multicolumn{3}{c}{CNN $\rightarrow$ Reddit} \\
\cmidrule(lr){3-5}
Type & Model & R1 (std) & R2 (std) & RL (std) \\
\midrule
\modelext & \ext & 24.30 (0.20) & 5.78 (0.08) & 19.87 (0.11) \\
\modelcmp & CUPS & 25.01 (0.15) & 5.96 (0.08) & 20.10 (0.09) \\
\bottomrule
\end{tabular}
\caption{Results on CNN $\rightarrow$ Reddit, reporting ROUGE with standard deviation across 5 independent runs with a random batch of 500 samples.}
\label{tab:appendix-ood-cnnreddit-var}
\end{table*}

\begin{table*}[t]
\setlength{\tabcolsep}{4pt}
\small
\centering
\begin{tabular}{clrrr}
\toprule
& & \multicolumn{3}{c}{XSum $\rightarrow$ WikiHow} \\
\cmidrule(lr){3-5}
Type & Model & R1 (std) & R2 (std) & RL (std) \\
\midrule
\modelext & \ext & 30.22 (0.05) & 8.43 (0.03) & 28.30 (0.03) \\
\modelcmp & CUPS & 30.52 (0.06) & 8.44 (0.01) & 28.48 (0.04) \\
\bottomrule
\end{tabular}
\caption{Results on XSum $\rightarrow$ WikiHow, reporting ROUGE with standard deviation across 5 independent runs with a random batch of 500 samples.}
\label{tab:appendix-ood-xsumwikihow-var}
\end{table*}

\begin{table*}[t]
\setlength{\tabcolsep}{4pt}
\small
\centering
\begin{tabular}{clrrrrrrrrr}
\toprule
 &  & \multicolumn{3}{c}{NYT $\rightarrow$ CNN} & \multicolumn{3}{c}{CNN $\rightarrow$ Reddit} & \multicolumn{3}{c}{XSum $\rightarrow$ WikiHow} \\
\cmidrule(lr){3-5} \cmidrule(lr){6-8} \cmidrule(lr){9-11}
Type & Model & R1 & R2 & RL & R1 & R2 & RL & R1 & R2 & RL \\
\midrule
\modelext & \ext & 31.90 & 13.04 & 28.42 & 23.76 & 5.66 & 18.95 & 29.44 & 8.25 & 27.41 \\
\modelcmp & CUPS & \textbf{33.98} & \textbf{13.25} & 30.39 & \textbf{25.01} & \textbf{5.96} & \textbf{20.10} & \textbf{30.52} & \textbf{8.44} & \textbf{28.48} \\
 & \quad - Salience & 33.74 & 13.19 & \textbf{30.46} & 24.30 & 5.78 & 19.87 & 30.22 & 8.43 & 28.30 \\
\bottomrule
\end{tabular}
\caption{Results on NYT $\rightarrow$ CNN, CNN $\rightarrow$ Reddit, and XSum $\rightarrow$ WikiHow after removing the salience model.}
\label{tab:appendix-ood-ablations}
\end{table*}

\section{Reproducibility}

Table~\ref{tab:appendix-dev-results} shows system results on the development sets of CNN/DM, CNN, WikiHow, XSum, and Reddit to aid the reproducibility of our system; both \ext and CUPS are included. Furthermore, in Table~\ref{tab:appendix-train-time}, we report several metrics to aid the training of the extraction and compression models. These specific metrics recorded by training models on a 32GB NVIDIA V100 GPU with the hyperparameters listed in Table~\ref{tab:appendix-training-hyperparams}.

\begin{table*}[t]
\setlength{\tabcolsep}{3pt}
\small
\centering
\begin{tabular}{clrrrrrrrrrrrrrrr}
\toprule
 &  & \multicolumn{3}{c}{CNN/DM} & \multicolumn{3}{c}{CNN} & \multicolumn{3}{c}{WikiHow} & \multicolumn{3}{c}{XSum} & \multicolumn{3}{c}{Reddit} \\
\cmidrule(lr){3-5} \cmidrule(lr){6-8} \cmidrule(lr){9-11} \cmidrule(lr){12-14} \cmidrule(lr){15-17}
Type & Model & R1 & R2 & RL & R1 & R2 & RL & R1 & R2 & RL & R1 & R2 & RL & R1 & R2 & RL \\
\midrule
\multicolumn{14}{l}{\textbf{Encoder: BERT}} \\
\midrule
\modelext & \ext & 43.37 & 20.50 & 39.86 & 31.85 & 12.98 & 28.53 & 30.20 & 8.58 & 28.07 & 23.67 & 4.52 & 17.89 & 24.20 & 5.78 & 18.77 \\
\modelcmp & CUPS & \textbf{43.68} & \textbf{20.51} & \textbf{40.16} & \textbf{34.26} & \textbf{13.63} & \textbf{30.93} & \textbf{31.55} & \textbf{8.95} & \textbf{29.42} & \textbf{25.37} & \textbf{4.93} & \textbf{19.44} & \textbf{25.51} & \textbf{6.17} & \textbf{19.96} \\
\midrule
\multicolumn{14}{l}{\textbf{Encoder: ELECTRA}} \\
\midrule
\modelext & \ext & 43.97 & 21.03 & 40.45 & 32.50 & 13.40 & 29.09 & 30.75 & 8.90 & 28.57 & 24.44 & 5.03 & 18.48 & 25.09 & 6.40 & 19.42 \\
\modelcmp & CUPS & \textbf{44.35} & \textbf{21.07} & \textbf{40.81} & \textbf{34.87} & \textbf{13.89} & \textbf{31.35} & \textbf{32.20} & \textbf{9.34} & \textbf{30.01} & \textbf{26.24} & \textbf{5.47} & \textbf{20.06} & \textbf{26.73} & \textbf{6.90} & \textbf{20.84} \\
\bottomrule
\end{tabular}
\caption{Results on the development sets of CNN/DM, CNN, WikiHow, XSum, and Reddit using the default CUPS system, leveraging both BERT$_\textrm{BASE}$ and ELECTRA$_\textrm{BASE}$ pre-trained encoders.}
\label{tab:appendix-dev-results}
\end{table*}

\begin{table*}[t]
\setlength{\tabcolsep}{4pt}
\small
\centering
\begin{tabular}{lrrrrrrr}
\toprule
Metrics & CNN/DM & CNN & NYT & WikiHow & XSum & Reddit & Google \\
\midrule
\multicolumn{8}{l}{\textbf{Model: Extraction}} \\
\midrule
Train Steps & 22K & 15K & 18K & 23K & 24K & 10K & --- \\
Time Elapsed (hrs/min) & 6h 48m & 3h 4m & 5h 52m & 5h 5m & 6h 6m & 1h 59m & --- \\
\midrule
\multicolumn{8}{l}{\textbf{Model: Compression}} \\
\midrule
Train Steps & 26K & 13K & 19K & 25K & 25K & 10K & 20K \\
Time Elapsed (hrs/min) & 3h 32m & 1h 27m & 2h 38m & 3h 26m & 3h 38m & 0h 56m & 1h 59m \\
\bottomrule
\end{tabular}
\caption{Number of training steps and total time elapsed for training extraction and compression models on CNN/DM, CNN, NYT, WikiHow, XSum, Reddit, and Google*. Models are benchmarked on a 32GB NVIDIA V100 GPU. *Google refers to the sentence compression dataset released by \citet{filippova-2013-overcoming}, which is only used to train the plausibility compression model.}
\label{tab:appendix-train-time}
\end{table*}

\end{document}